\documentclass[sigconf]{acmart}
\AtBeginDocument{%
 }

\copyrightyear{2024}
\acmYear{2024}
\setcopyright{rightsretained}
\acmConference[WWW '24 Companion]{Companion Proceedings of the ACM Web Conference 2024}{May 13--17, 2024}{Singapore, Singapore}
\acmBooktitle{Companion Proceedings of the ACM Web Conference 2024 (WWW '24 Companion), May 13--17, 2024, Singapore, Singapore}
\acmDOI{10.1145/3589335.3651978 }
\acmISBN{979-8-4007-0172-6/ 24/05}

\makeatletter
\gdef\@copyrightpermission{
  \begin{minipage}{0.3\columnwidth}
   \href{https://creativecommons.org/licenses/by/4.0/}{\includegraphics[width=0.90\textwidth]{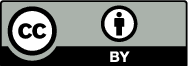}}
  \end{minipage}\hfill
  \begin{minipage}{0.7\columnwidth}
   \href{https://creativecommons.org/licenses/by/4.0/}{This work is licensed under a Creative Commons Attribution International 4.0 License.}
  \end{minipage}
  \vspace{5pt}
}
\makeatother

\usepackage{algorithm}
\usepackage{algorithmic}
\usepackage{xspace}
\usepackage{bbm}
\usepackage{balance}

\newcommand{\entities}{\ensuremath{\mathcal{E}}\xspace}
\newcommand{\relations}{\ensuremath{\mathcal{R}}\xspace}

\newcommand{\kg}{\ensuremath{\mathcal{G}}\xspace}
\newcommand{\kge}{\phi}

\begin{document}

\title{Universal Knowledge Graph Embeddings}

\settopmatter{authorsperrow=4}
\author{N'Dah Jean Kouagou}
\email{ndah.jean.kouagou@upb.de}
\orcid{0000-0002-4217-897X}
\affiliation{%
  \institution{Department of Computer Science, Paderborn University}
  \city{Paderborn}
  \state{NRW}
  \country{Germany}
}

\author{Caglar Demir}
\email{	caglar.demir@upb.de}
\orcid{0000-0001-8970-3850}
\affiliation{%
  \institution{Department of Computer Science, Paderborn University}
  \city{Paderborn}
  \state{NRW}
  \country{Germany}
}

\author{Hamada M. Zahera}
\email{hamada.zahera@upb.de}
\orcid{0000-0003-0215-1278}
\affiliation{%
  \institution{Department of Computer Science, Paderborn University}
  \city{Paderborn}
  \state{NRW}
  \country{Germany}
  \postcode{33098}
}

\author{Adrian Wilke}
\email{adrian.wilke@upb.de}
\orcid{0000-0002-6575-807X}
\affiliation{%
  \institution{Department of Computer Science, Paderborn University}
  \city{Paderborn}
  \state{NRW}
  \country{Germany}
}

\author{Stefan Heindorf}
\email{heindorf@upb.de}
\orcid{0000-0002-4525-6865}
\affiliation{%
  \institution{Department of Computer Science, Paderborn University}
  \city{Paderborn}
  \state{NRW}
  \country{Germany}
}

\author{Jiayi Li}
\email{jiayili@mail.uni-paderborn.de}
\orcid{0009-0001-9475-8159}
\affiliation{%
  \institution{Department of Computer Science, Paderborn University}
  \city{Paderborn}
  \state{NRW}
  \country{Germany}
}

\author{Axel-Cyrille Ngonga Ngomo}
\email{axel.ngonga@upb.de}
\orcid{0000-0001-7112-3516}
\affiliation{%
  \institution{Department of Computer Science, Paderborn University}
  \city{Paderborn}
  \state{NRW}
  \country{Germany}
}

\renewcommand{\shortauthors}{N’Dah Jean Kouagou et al.}

\begin{abstract}
  A variety of knowledge graph embedding approaches have been developed.
Most of them obtain embeddings by learning the structure of the knowledge graph within a link prediction setting.
As a result, the embeddings reflect only the structure of a single knowledge graph, and embeddings for different knowledge graphs are not aligned, e.g., they cannot be used to find similar entities across knowledge graphs via nearest neighbor search. However, knowledge graph embedding applications such as \textit{entity disambiguation} require a more global representation, i.e., a representation that is valid across multiple sources.
We propose to learn \textit{universal knowledge graph embeddings} from large-scale interlinked knowledge sources. To this end, we fuse large knowledge graphs based on the \texttt{owl:sameAs} relation such that every entity is represented by a unique identity. We instantiate our idea by computing universal embeddings based on DBpedia and Wikidata yielding embeddings for about 180 million entities, 15 thousand relations, and 1.2 billion triples. We believe our computed embeddings will support the emerging field of graph foundation models. Moreover, we develop a convenient API to provide embeddings as a service. Experiments on link prediction suggest that \textit{universal knowledge graph embeddings} encode better semantics compared to embeddings computed on a single knowledge graph. For reproducibility purposes, we provide our source code and datasets open access.\footnote{\url{https://github.com/dice-group/Universal_Embeddings}}
\end{abstract}


\maketitle

\section{Introduction}

Knowledge graph embedding (KGE) models are used for a multitude of applications~\cite{Wang2017Knowledge,dai2020survey}, including link prediction, triple classification, entity classification, and entity resolution. Several approaches to compute KGEs have been proposed in the literature, e.g., TransE~\cite{Bordes2013Translating}, RDF2Vec~\cite{Ristoski2016RDF2Vec}, and ComplEx~\cite{trouillon2016complex}.

While pretrained models for few knowledge graphs (KGs) are available, their embedding spaces are not aligned, i.e., same entities have different representations across different knowledge graphs. As a result, the usability of such embeddings is often limited to downstream tasks on the KG they were trained on \cite{jain2021embeddings}. 
However, a growing number of real-world applications of KG embeddings (e.g., graph foundation models~\cite{liu2023towards,galkin2023towards}) require entities to have a representation that integrates information from multiple sources.

The need for these unified representations for entities recently motivated several works \cite{Chen2017multilingual,sun2017cross,Sun2018Bootstrapping,Wang2018GCNAlign,Zhang2019Multi-view,Trisedya2019Entity}. Some of these approaches are tailored towards multi-lingual KG embeddings, i.e., the task of computing aligned embeddings between multiple language versions of the same KG by relying on the available \texttt{owl:sameAs} links~\cite{Chen2017multilingual,sun2017cross,Wang2018GCNAlign}. Other approaches employ a bootstrapping strategy based on the matching scores between entities during training~\cite{Sun2018Bootstrapping} or use additional information on entities such as attribute embeddings~\cite{Trisedya2019Entity,Zhang2019Multi-view}.
Although current alignment approaches for KGs have shown promising results on benchmark datasets, they inherently suffer from scalability issues. This is corroborated by the lack of pretrained KG embeddings for large datasets such as Wikidata~\cite{vrandevcic2014wikidata} and DBpedia~\cite{auer2007dbpedia} using the aforementioned approaches. Moreover, most entity alignment approaches can only handle two KGs at a time, and do not assign the same embedding vector to matching entities\footnote{Two entities are matching if they correspond to the same real world entity}.

In this paper, we merge a given set of KGs into a single KG to compute embeddings that capture comprehensive information about each entity. We call these embeddings \emph{universal knowledge graph embeddings}. By assigning a unique ID to all matching entities, we not only reduce memory consumption and computation costs but also tackle KG incompleteness\textemdash a well-known issue in the research community~\cite{zhao2022improving,wang2019learning,zhang2023variational}. For instance, Wikidata and DBpedia contain 360 and 138 triples about Iraq, respectively. Consequently, a traditional KGE model trained on either KG can only capture incomplete information about their shared entities. Our approach mitigates this limitation by integrating information from both KGs into the embeddings.

To quantify the quality of our embeddings, we apply our approach to 4 KGE models, i.e., DistMult\cite{yang2015embedding}, ComplEx\cite{trouillon2016complex}, QMult~\cite{demir2020convolutional}, and ConEx~\cite{demir2021convolutional}, and evaluate their performance on link prediction. Overall, our results suggest that the benefits of using additional information derived from \texttt{sameAs} links become particularly noticeable in ConEx. We use the latter to compute and provide high-quality unified embeddings for the most populous KGs of the Linked Open Data (LOD) cloud\footnote{LOD: \url{https://lod-cloud.net}}, i.e., DBpedia and Wikidata. The merged graph encompasses about 180 million entities, 15 thousand relations, and 1.2 billion triples. Moreover, we develop an API\footnote{API: \url{https://embeddings.cc/}} with convenient methods to make the computed embeddings easily accessible. 

\section{Related Work} \label{sec:relatedwork}
A knowledge graph embedding (KGE) model denoted by $\kge$
maps a knowledge graph (KG) into a continuous vector space---commonly by solving an optimization problem. This optimization aims to preserve the structural information of the input KG.
For example, transitional distance models such as TransE~\cite{Bordes2013Translating} compute embeddings by modelling each triple as a translation between its head and tail entities. Specifically, TransE represents both entities and relations as vectors in the same semantic space and learns embeddings by minimizing the distance between $h+r$ and $t$ for every triple $(h,r,t)$ in the considered KG. DistMult~\cite{yang2015embedding} adopts the scoring technique of TransE but uses multiplications to model entiy and relation interactions.
Other types of KG embedding models include ComplEx \cite{trouillon2016complex}, ConEx \cite{demir2020convolutional} and RESCAL \cite{nickel2011three}. ComplEx and ConEx model entities and relations as complex vectors (i.e. with real and imaginary parts) to handle both symmetric and antisymmetric relations. As ComplEx cannot handle transitive relations (see \citeauthor{sun2019rotate}~\citeyear{sun2019rotate}), ConEx further improves on ComplEx by applying a 2D convolution operation on complex-valued embeddings of head entities and relations. By associating each relation with a matrix, RESCAL captures pairwise interactions between entities and is regarded as one of the most expressive models~\cite{Wang2017Knowledge}. As KGs grow in size, computation-efficient algorithms are required to train KGs consisting of millions of entities and billions of triples. \citeauthor{zheng2020dgl}~\citeyear{zheng2020dgl} developed DGL-KE, an open-source package that employs several optimization techniques to accelerate training on large KGs. For example, they partition a large KG to perform gradient updates on each partition and regularly fetch embeddings from other partitions which involves a significant communication overhead.

Although KG embeddings can benefit downstream tasks such as link prediction and KG completion, their successful application is often limited to the KGs they were trained on. As a result, applications to other tasks, such as \emph{entity resolution on two or multiple KGs}, require aligned KG embeddings.

\begin{algorithm}[tb]
	\caption{Function \textsc{FuseKGs}}
	\label{alg:fuse}
	\begin{algorithmic}[1]
	    \REQUIRE $\kg_1$, $\kg_2$, $\dots$, $\kg_N$ 
	    \ENSURE $\kg^*$  \quad \# \textit{merged knowledge graphs}
		\STATE $\kg^* \leftarrow \kg_1$ \quad \# \textit{Initialization}
		\FOR{$i = 2, \dots, N$}
		\FOR{($e_1$, $r$, $e_2$) $\in$ $\kg_i$}
		\IF{$\exists\ e^*_1 \in \entities_{\kg^*}$ such that $e_1\texttt{:sameAs:}e^*_1$}
		\STATE $e_1 \leftarrow e^*_1$ \quad \# \textit{Rename entity $e_1$}
		\ENDIF
		\IF{$\exists\ e^*_2 \in \entities_{\kg^*}$ such that $e_2\texttt{:sameAs:}e^*_2$}
		\STATE $e_2 \leftarrow e^*_2$ \quad \# \textit{Rename entity $e_2$}
		\ENDIF
		\STATE Add ($e_1$, $r$, $e_2$) to $\kg^*$
		\ENDFOR
		\ENDFOR
		\STATE \textbf{return} $\kg^*$
	\end{algorithmic}
\end{algorithm}

\section{Universal Knowledge Graph Embeddings}
\label{sec:universalkge}

\subsection{Preliminaries}
A KG \kg can be regarded as a set of triples $\kg= \{(e_1^{(i)},r^{(i)},e_2^{(i)})\}_{i=1}^{n}$ $\subseteq \entities_{\kg} \times \relations_{\kg} \times \entities_{\kg}$, where $\entities_{\kg}$ and $\relations_{\kg}$ represent its sets of entities and relations, respectively. When there is no ambiguity, we simply write \entities and \relations. Let $\kg_1,\dots,\kg_N$ denote $N$ KGs (in an arbitrary order), e.g., $\text{DBpedia}, \text{Wikidata}, \text{Freebase}$. Alignments between $\kg_i$ and $\kg_j$ are given by \texttt{sameAs} links. We use these links to fuse the given KGs as described in the next section.

\subsection{Graph Fusion and Embedding Computation}
In this work, we fuse all KGs $\kg_i$ for $i=1, \dots, N$ into a single KG $\kg^*$ where all aligned entities are represented by a unique ID. Algorithm \ref{alg:fuse} describes how the fusion is carried out. First, the algorithm chooses a reference KG%
\footnote{It does not matter which KG is chosen as we end up with the same number of triples in any case.} (in this work, we select $\kg_1$) as the initial set of triples for $\kg^*$ (line 1). Then, it iterates over the rest of the KGs (line 2) and adds their triples (line 10). In this process, entities that are already present in $\kg^*$ via \texttt{sameAs} links are renamed accordingly (lines 3--9).
Once $\kg^*$ is constructed, a KGE model (e.g. ConEx) can be applied to learn universal embeddings.

\begin{table}[tb]
	\centering
	\caption{Statistics of the full datasets for \emph{universal knowledge graph embeddings}. \texttt{Deg.} denotes the average degree of entities.}
	\label{tab:full-data}
	\small
	\setlength{\tabcolsep}{6pt}
	\begin{tabular}{@{}lccccc@{}}
		\toprule
		Dataset & $| \entities|$ & $|\relations|$ & $|\kg|$ & $|\texttt{sameAs}|$ & \texttt{Deg.}\\
		\midrule
		DBpedia & 91,684,304 & 13,783 & \phantom{0}616,564,603 & 33,860,047 & 13.45\\
		Wikidata & 94,468,182 &  \phantom{0}1,436 & \phantom{0}667,666,110 & 33,860,047 & 14.14\\
		MERGE & 179,706,494 & 15,219 &1,284,230,713 & 33,860,047 & 14.29\\
		\bottomrule
	\end{tabular}
\end{table}

\subsection{Knowledge Graphs}
We downloaded and preprocessed the September 2022 version of DBpedia~\cite{auer2007dbpedia} and the March 2022 version of Wikidata~\cite{vrandevcic2014wikidata}. In this work, we only consider the English version of DBpedia, and its external links to Wikidata, i.e., \texttt{sameAs} links. The preprocessing step is concerned with the removal of non triplet-formatted files and literals.

\paragraph{DBpedia ($\kg_1$).}
DBpedia\footnote{\url{https://databus.dbpedia.org/dbpedia/collections/dbpedia-snapshot-2022-09/}} is the most popular and prominent KG in the LOD. It is automatically created based on Wikipedia information, such as infobox tables, categorizations, and links to external websites. Since DBpedia serves as the hub for LOD, it contains many links to other LOD datasets such as Freebase, Caligraph, and Wikidata. 

\paragraph{Wikidata ($\kg_2$).}
Wikidata\footnote{\url{https://dumps.wikimedia.org/wikidatawiki/entities/}} is a community-created knowledge base providing factual information to Wikipedia and other projects by the Wikimedia Foundation. As of April 2022, Wikidata contains over $97$ million items and $1.37$ billion statements. Each item page contains labels, short descriptions, aliases, statements, and site links. Each statement consists of a claim and an optional reference, and each claim consists of a property-value pair and optional qualifiers.

\paragraph{Statistics of Knowledge Graphs.} Table~\ref{tab:full-data} presents the statistics of DBpedia and Wikidata after our preprocessing step. MERGE is obtained by applying Algorithm \ref{alg:fuse} to \{DBpedia, Wikidata\}. In a normal scenario, the sum of the numbers of entities of DBpedia and Wikidata should be equal to that of MERGE and \texttt{sameAs} links. However, some entities in  DBpedia were matched with multiple entities in Wikidata via \texttt{sameAs} links, and vice versa. This caused the equality not to hold as can be seen in the table.

\section{Experiments}
\label{sec:experiments}
We conduct our experiments to answer one fundamental question: \textit{``How do our universal knowledge embeddings compare to embeddings from traditional KGE approaches?''}
To this end, we set up a link prediction task where we compare two independently trained instances of the same embedding model (see the next sections for more details).

\begin{table}[tb]
	\small
	\centering
	\setlength{\tabcolsep}{8.5pt}
	\caption{Statistics of evaluation datasets. \texttt{Deg}. is the average degree of entities. } 
	\label{tab:data}
	\begin{tabular}{@{}lccccc@{}}
		\toprule
		Dataset & $|\entities|$ & $|\relations|$ & $|\kg|$ & $|\texttt{sameAs}|$ & \texttt{Deg.}\\
		\midrule
		DBpedia$_{train}$ & 31,116 & \phantom{0}~392 & \phantom{0}69,667 & 22,102 &4.48\\
		DBpedia$_{test}$ & 15,602 & \phantom{0}~279 & \phantom{0}15,374 & 10,471 & 1.97\\
		Wikidata$_{train}$ & 72,058 &  \phantom{0}~707 & 235,814 & 22,102 & 6.55\\
		Wikidata$_{test}$ & 41,137 &  \phantom{0}~465 & \phantom{0}53,761 & 10,471 & 2.61\\
		MERGE$_{train}$ & 81,836 &  1,099 & 305,481 & 22,102 & 7.47\\
		\bottomrule
	\end{tabular}%
\end{table}

\subsection{Evaluation Setup}
\label{subsec:experimental-setup}
\paragraph{Evaluation Datasets.}
We conduct our experiments on subsets of DBpedia and Wikidata, due to the computational complexity of our evaluation metrics. Specifically, we randomly select 1\% of entities in DBpedia that share \texttt{sameAs} links to Wikidata, then we obtain their 1-hop neighborhood together with the corresponding relation types. We then analogously compute the corresponding subset of Wikidata by using entities identified by the 1\% initially selected in DBpedia. The samples we obtain are then randomly split into training and test datasets. Overall, we obtain five datasets for our experiments: DBpedia$_{train}$ and DBpedia$_{test}$ (split of the DBpedia sample), Wikidata$_{train}$ and Wikidata$_{test}$ (split of the Wikidata sample), and MERGE$_{train}$ (merge of DBpedia$_{train}$ and Wikidata$_{train}$ using Algorithm \ref{alg:fuse}). For the sake of clarity, we use the notations DBpedia+ and Wikidata+ to refer to MERGE$_{train}$ depending on whether we are evaluating embedding models on DBpedia or Wikidata.
Note that the splits are performed in a way that all entities and relation types in training datasets also appear in the test datasets. This allows us to not encounter out-of-vocabulary entities and relations at inference time.  
The sizes of the splits are specified in Table~\ref{tab:data}. The average degree abbreviated as \texttt{Deg.} represents the average number of edges connected to an entity.

\paragraph{Metrics.}
We use two standard metrics to evaluate KGEs: \textit{hits}@$k$ (H@$k$) and \textit{mean reciprocal rank} (\textsc{MRR}).
Formally, let \kg be a knowledge graph, i.e., a set of triples. We denote by $\mathit{rank}[e_1|r,e_2]$ the rank of the score of $e_1$ given the relation $r$ and the tail entity $e_2$ among the set of all scores $\{\mathit{score}(e|r,e_2)\text{ s. t. }e \in \entities_{\kg}\}$. Similarly, $\mathit{rank}[e_2|e_1,r]$ denotes the rank of the score of $e_2$ given the head entity $e_1$ and the relation $r$ among $\{\mathit{score}(e|e_1,r) \text{ s. t. } e \in \entities_{\kg}\}$. We define the metrics Hits@$k$ and \textsc{MRR} as

{
\small
\begin{align}
\text{Hits@}k =& \frac{1}{2|\kg|} \sum_{(e_1, r, e_2)\in \kg}\mathbbm{1}(\mathit{rank}[e_1|r,e_2]\le k) \nonumber\\ +& \mathbbm{1}(\mathit{rank}[e_2|e_1,r]\le k),\\
\textsc{MRR} =& \frac{1}{2|\kg|} \sum_{(e_1, r, e_2)\in \kg}\frac{1}{\mathit{rank}[e_1|r,e_2]}\nonumber\\
+& \frac{1}{\mathit{rank}[e_2|e_1,r]}.
\end{align}
}

\paragraph{Hardware.}
The entire DBpedia and Wikidata datasets for which we provide embeddings as a service were processed on a virtual machine (VM) with 128 CPUs (AMD EPYC 7742 64-Core Processor) and 1TB RAM.
The computation of universal knowledge graph embeddings was carried out using the DICE embedding framework~\cite{demir2022hardware} on a 1TB RAM VM with 2 NVIDIA RTX A5000 GPUs of 24GB memory each.

\begin{table}[tb]
\caption{Link prediction results in terms of mean reciprocal rank (MRR) and hits@k (H@k). We compare the performance of each embedding model on the two types of datasets: single KG (DBpedia, Wikidata) and enriched KG (DBpedia+, Wikidata+). Hence, the bold values correspond to the best performance achieved row-wise.
All models use 32 embedding dimensions and have approximately the same number of parameters.}
\label{tab:results-d-32}
\centering
\small
\setlength{\tabcolsep}{2.6pt}
\begin{tabular}{@{}lcccccccc@{}}
\toprule
  &\multicolumn{4}{c}{DBpedia}&\multicolumn{4}{c}{DBpedia+}\\
  \cmidrule(lr){2-5} \cmidrule(l){6-9}
                        & MRR & H@1  &H@3   & H@10 & MRR  & H@1 &H@3  & H@10\\
  \midrule
ConEx-train             &0.371&0.298& 0.396&0.502 & \textbf{0.898}&\textbf{0.863}&\textbf{0.924}&\textbf{0.962}\\
ConEx-test              &0.181&0.134& 0.191&0.269 & \textbf{0.349}&\textbf{0.280}&\textbf{0.378}&\textbf{0.478}\\
\midrule
ComplEx-train             &0.181 &0.122 &0.205 &0.292  & \textbf{0.278}&\textbf{0.206} &\textbf{0.306} &\textbf{0.408}\\
ComplEx-test              &0.136 &0.103 &0.143 &0.194  & \textbf{0.163} &\textbf{0.123} &\textbf{0.170} &\textbf{0.237}\\
\midrule
QMult-train             &\textbf{0.230} &\textbf{0.168} &\textbf{0.250} &\textbf{0.348} & 0.129&0.099 &0.135 &0.183\\
QMult-test              &\textbf{0.148} &\textbf{0.112} &\textbf{0.155} &\textbf{0.213} & 0.064& 0.041& 0.070& 0.113\\
\midrule
DistMult-train             &\textbf{0.145} &\textbf{0.100} &\textbf{0.152} &\textbf{0.243} & 0.125& 0.094& 0.133&0.183\\
DistMult-test              &\textbf{0.118} &\textbf{0.084} &\textbf{0.125} &\textbf{0.177} & 0.057&0.032 &0.060 &0.109\\
\midrule
&\multicolumn{4}{c}{Wikidata}&\multicolumn{4}{c}{Wikidata+}\\
  \cmidrule(lr){2-5} \cmidrule(l){6-9}
                        & MRR & H@1  &H@3   & H@10 & MRR  & H@1 &H@3  & H@10\\
  \midrule
ConEx-train             &0.795&0.730&0.842& 0.911& \textbf{0.855}&\textbf{0.806}&\textbf{0.889}&\textbf{0.941}\\
ConEx-test              &0.280&0.227& 0.302&0.376 & \textbf{0.299}&\textbf{0.247}&\textbf{0.324}&\textbf{0.392}\\
\midrule
ComplEx-train             &0.232 &0.188 &0.244 &0.303 & \textbf{0.278}& \textbf{0.206}& \textbf{0.306}&\textbf{0.408}\\
ComplEx-test              &0.126 &0.088 &0.140 &0.197 & \textbf{0.166}&\textbf{0.123} &\textbf{0.181} &\textbf{0.247}\\
\midrule
QMult-train             &\textbf{0.138} &\textbf{0.105} &\textbf{0.147} &\textbf{0.199} & 0.129&0.099 &0.135 &0.183\\
QMult-test              &0.077 &0.053 &0.085 &0.122 & \textbf{0.092}& \textbf{0.069}& \textbf{0.100}&\textbf{0.135}\\
\midrule
DistMult-train             &\textbf{0.190} &\textbf{0.145} &\textbf{0.207} &\textbf{0.263} & 0.125& 0.094&0.133 &0.183\\
DistMult-test              & \textbf{0.107}& \textbf{0.075}& \textbf{0.118}&\textbf{0.165}& 0.086&0.060 &0.096 &0.136\\
\bottomrule
\end{tabular}
\end{table}

\subsection{Results and Discussion}
In Table~\ref{tab:results-d-32}, we present the results of our experiments comparing the performance of embedding models trained on a single KG against those trained on merged KGs (by leveraging \texttt{sameAs} links as described in Algorithm \ref{alg:fuse}). Four embedding models are considered: ConEx \cite{demir2020convolutional} , ComplEx \cite{trouillon2016complex}, QMult \cite{demir2021convolutional} and DistMult \cite{yang2015embedding}. 
From the table, we can observe that ConEx achieves the highest performance w.r.t. all metrics on all datasets. Moreover, its performance on the merged KGs (DBpedia+ and Wikidata+) is notably higher as compared to that on DBpedia and Wikidata. The ComplEx model also performs better on DBpedia+ and Wikidata+ than on DBpedia and Wikidata, respectively. On the other side, DistMult and QMult perform poorly on both the training and test datasets. 

One would expect that additional information added about entities in DBpedia+ and Wikidata+ improves the performance of embedding models on downstream tasks such as link prediction. 
Although this is clearly the case for ConEx and ComplEx (with up to 2$\times$ improvement for ConEx), we observed the opposite on DistMult and QMult. Interestingly, the poor-performing models correspond to the extreme cases of model complexity, i.e., DistMult is the simplest and QMult is the most expressive among the four models we considered. 
This suggests that with 32 embedding dimensions, DistMult cannot learn meaningful representations for entities and relations in our evaluation data due to its simplicity. Likewise, QMult fails to find optimal representations of entities and relations because it cannot encode its inherent high degree of freedom in 32 dimensions. The ConEx architecture appears to balance well between expressiveness and the chosen number of embedding dimensions. In fact, our preliminary experiments with 300 embedding dimensions ranked DistMult the top best-performing model ahead of ConEx and ComplEx at the cost of longer training times and memory consumption.
In view of this observation, we use ConEx with 32 embedding dimensions to compute our universal embeddings for large KGs and provide them on a platform (see next section).
The answer to the fundamental question behind our work is hence that \emph{we can learn rich embeddings on a KG that integrates information (about entities) from different external sources, in particular other KGs}. A precondition to achieve this goal is intrinsic to common challenges in representation learning: find fitting hyper-parameters. 

\section{Implementation of a Service Platform}
\label{sec:platform}
We offer computed data as an open service, following the FAIR principles \cite{wilkinson2016fair}, so that the universal embeddings are available to a broad audience. The platform\footnote{\url{https://embeddings.cc/}} consists of a RESTful, TLS-secured API along with a website and an interactive documentation.
The API contains a hidden webservice for developers to maintain data and a public webservice with eight methods providing RDF entity identifiers and related embeddings.

Regarding the FAIR principles, the data is \emph{findable} as existing entity identifiers are used and can be accessed and explored with an autocomplete feature on the website.
The big data amount is \emph{accessible} as data subsets can be retrieved using API methods such as \textit{random} or \textit{autocomplete}, which explore the full data. The accessibility is additionally enhanced by meta queries like the size of the offered datasets. \emph{Interoperability} is given by reusing existing RDF namespaces and identifiers.
In addition, the API is versioned and uses the lightweight JSON format in Python and JavaScript.

\section{Conclusion and Outlook}
\label{sec:conclusion}
In this paper, we discuss the challenges related to computing embeddings for entities shared across multiple knowledge graphs. In particular, we note the lack of such embeddings for large knowledge graphs and propose a simple but effective approach to compute embeddings for shared entities. Given a countable set of knowledge graphs, our approach iterates over all triples and assigns a unique ID to all matching entities (i.e. shared entities). An embedding model is then applied to learn embeddings on the resulting graph\textemdash our embeddings are called universal knowledge graph embeddings. We use our approach to compute embeddings for recent versions of DBpedia and Wikidata, and provide them as an open service via a convenient API. Experiments on link prediction suggest that our universal embeddings are better than those computed on separate knowledge graphs. Regarding the API, we currently provide embeddings via autocomplete search and random entity selection. In future releases, we will integrate an approximation of embedding-level nearest neighbour search to support real-time queries of similar entities over the complete data. We will also collect more large-scale knowledge graphs from the Linked Open Data Cloud to update our universal embeddings.

\section*{Acknowledgements}
This work has received funding from the European Union's Horizon 2020 research and innovation programme within the project KnowGraphs under the Marie Skłodowska-Curie grant No 860801, the European Union’s Horizon Europe research and innovation programme within the project ENEXA under the grant No 101070305, and the European Union’s Horizon Europe research and innovation programme within the project LEMUR under the Marie Skłodowska-Curie grant agreement No 101073307.
This work has also been supported by the Ministry of Culture and Science of North Rhine-Westphalia (MKW NRW) within the project SAIL under the grant No NW21-059D and by the Deutsche Forschungsgemeinschaft (DFG, German Research Foundation): TRR 318/1 2021~–~438445824.

\bibliographystyle{ACM-Reference-Format}
\balance
\bibliography{bibliography}


\begin{thebibliography}{27}


\ifx \showCODEN    \undefined \def \showCODEN     #1{\unskip}     \fi
\ifx \showDOI      \undefined \def \showDOI       #1{#1}\fi
\ifx \showISBNx    \undefined \def \showISBNx     #1{\unskip}     \fi
\ifx \showISBNxiii \undefined \def \showISBNxiii  #1{\unskip}     \fi
\ifx \showISSN     \undefined \def \showISSN      #1{\unskip}     \fi
\ifx \showLCCN     \undefined \def \showLCCN      #1{\unskip}     \fi
\ifx \shownote     \undefined \def \shownote      #1{#1}          \fi
\ifx \showarticletitle \undefined \def \showarticletitle #1{#1}   \fi
\ifx \showURL      \undefined \def \showURL       {\relax}        \fi
\providecommand\bibfield[2]{#2}
\providecommand\bibinfo[2]{#2}
\providecommand\natexlab[1]{#1}
\providecommand\showeprint[2][]{arXiv:#2}

\bibitem[Auer et~al\mbox{.}(2007)]%
        {auer2007dbpedia}
\bibfield{author}{\bibinfo{person}{S{\"{o}}ren Auer},
  \bibinfo{person}{Christian Bizer}, \bibinfo{person}{Georgi Kobilarov},
  \bibinfo{person}{Jens Lehmann}, \bibinfo{person}{Richard Cyganiak}, {and}
  \bibinfo{person}{Zachary~G. Ives}.} \bibinfo{year}{2007}\natexlab{}.
\newblock \showarticletitle{DBpedia: {A} Nucleus for a Web of Open Data}. In
  \bibinfo{booktitle}{\emph{{ISWC/ASWC}}}, Vol.~\bibinfo{volume}{4825}.
  \bibinfo{publisher}{Springer}, \bibinfo{pages}{722--735}.
\newblock


\bibitem[Bordes et~al\mbox{.}(2013)]%
        {Bordes2013Translating}
\bibfield{author}{\bibinfo{person}{Antoine Bordes}, \bibinfo{person}{Nicolas
  Usunier}, \bibinfo{person}{Alberto Garc{\'{\i}}a{-}Dur{\'{a}}n},
  \bibinfo{person}{Jason Weston}, {and} \bibinfo{person}{Oksana Yakhnenko}.}
  \bibinfo{year}{2013}\natexlab{}.
\newblock \showarticletitle{Translating Embeddings for Modeling
  Multi-relational Data}. In \bibinfo{booktitle}{\emph{{NIPS}}}.
  \bibinfo{pages}{2787--2795}.
\newblock


\bibitem[Chen et~al\mbox{.}(2017)]%
        {Chen2017multilingual}
\bibfield{author}{\bibinfo{person}{Muhao Chen}, \bibinfo{person}{Yingtao Tian},
  \bibinfo{person}{Mohan Yang}, {and} \bibinfo{person}{Carlo Zaniolo}.}
  \bibinfo{year}{2017}\natexlab{}.
\newblock \showarticletitle{Multilingual Knowledge Graph Embeddings for
  Cross-lingual Knowledge Alignment}. In \bibinfo{booktitle}{\emph{{IJCAI}}}.
  \bibinfo{pages}{1511--1517}.
\newblock


\bibitem[Dai et~al\mbox{.}(2020)]%
        {dai2020survey}
\bibfield{author}{\bibinfo{person}{Yuanfei Dai}, \bibinfo{person}{Shiping
  Wang}, \bibinfo{person}{Neal~N Xiong}, {and} \bibinfo{person}{Wenzhong Guo}.}
  \bibinfo{year}{2020}\natexlab{}.
\newblock \showarticletitle{A survey on knowledge graph embedding: Approaches,
  applications and benchmarks}.
\newblock \bibinfo{journal}{\emph{Electronics}} \bibinfo{volume}{9},
  \bibinfo{number}{5} (\bibinfo{year}{2020}), \bibinfo{pages}{750}.
\newblock


\bibitem[Demir et~al\mbox{.}(2021)]%
        {demir2021convolutional}
\bibfield{author}{\bibinfo{person}{Caglar Demir}, \bibinfo{person}{Diego
  Moussallem}, \bibinfo{person}{Stefan Heindorf}, {and}
  \bibinfo{person}{Axel{-}Cyrille~Ngonga Ngomo}.}
  \bibinfo{year}{2021}\natexlab{}.
\newblock \showarticletitle{Convolutional Hypercomplex Embeddings for Link
  Prediction}. In \bibinfo{booktitle}{\emph{{ACML}}}
  \emph{(\bibinfo{series}{Proceedings of Machine Learning Research},
  Vol.~\bibinfo{volume}{157})}. \bibinfo{publisher}{{PMLR}},
  \bibinfo{pages}{656--671}.
\newblock


\bibitem[Demir and Ngomo(2021)]%
        {demir2020convolutional}
\bibfield{author}{\bibinfo{person}{Caglar Demir} {and}
  \bibinfo{person}{Axel{-}Cyrille~Ngonga Ngomo}.}
  \bibinfo{year}{2021}\natexlab{}.
\newblock \showarticletitle{Convolutional Complex Knowledge Graph Embeddings}.
  In \bibinfo{booktitle}{\emph{{ESWC}}}, Vol.~\bibinfo{volume}{12731}.
  \bibinfo{publisher}{Springer}, \bibinfo{pages}{409--424}.
\newblock


\bibitem[Demir and Ngomo(2022)]%
        {demir2022hardware}
\bibfield{author}{\bibinfo{person}{Caglar Demir} {and}
  \bibinfo{person}{Axel{-}Cyrille~Ngonga Ngomo}.}
  \bibinfo{year}{2022}\natexlab{}.
\newblock \showarticletitle{Hardware-agnostic computation for large-scale
  knowledge graph embeddings}.
\newblock \bibinfo{journal}{\emph{Softw. Impacts}}  \bibinfo{volume}{13}
  (\bibinfo{year}{2022}), \bibinfo{pages}{100377}.
\newblock


\bibitem[Galkin et~al\mbox{.}(2023)]%
        {galkin2023towards}
\bibfield{author}{\bibinfo{person}{Mikhail Galkin}, \bibinfo{person}{Xinyu
  Yuan}, \bibinfo{person}{Hesham Mostafa}, \bibinfo{person}{Jian Tang}, {and}
  \bibinfo{person}{Zhaocheng Zhu}.} \bibinfo{year}{2023}\natexlab{}.
\newblock \showarticletitle{Towards Foundation Models for Knowledge Graph
  Reasoning}.
\newblock \bibinfo{journal}{\emph{CoRR}}  \bibinfo{volume}{abs/2310.04562}
  (\bibinfo{year}{2023}).
\newblock


\bibitem[Jain et~al\mbox{.}(2021)]%
        {jain2021embeddings}
\bibfield{author}{\bibinfo{person}{Nitisha Jain},
  \bibinfo{person}{Jan{-}Christoph Kalo}, \bibinfo{person}{Wolf{-}Tilo Balke},
  {and} \bibinfo{person}{Ralf Krestel}.} \bibinfo{year}{2021}\natexlab{}.
\newblock \showarticletitle{Do Embeddings Actually Capture Knowledge Graph
  Semantics?}. In \bibinfo{booktitle}{\emph{{ESWC}}},
  Vol.~\bibinfo{volume}{12731}. \bibinfo{publisher}{Springer},
  \bibinfo{pages}{143--159}.
\newblock


\bibitem[Liu et~al\mbox{.}(2023)]%
        {liu2023towards}
\bibfield{author}{\bibinfo{person}{Jiawei Liu}, \bibinfo{person}{Cheng Yang},
  \bibinfo{person}{Zhiyuan Lu}, \bibinfo{person}{Junze Chen},
  \bibinfo{person}{Yibo Li}, \bibinfo{person}{Mengmei Zhang},
  \bibinfo{person}{Ting Bai}, \bibinfo{person}{Yuan Fang},
  \bibinfo{person}{Lichao Sun}, \bibinfo{person}{Philip~S. Yu}, {and}
  \bibinfo{person}{Chuan Shi}.} \bibinfo{year}{2023}\natexlab{}.
\newblock \showarticletitle{Towards Graph Foundation Models: {A} Survey and
  Beyond}.
\newblock \bibinfo{journal}{\emph{CoRR}}  \bibinfo{volume}{abs/2310.11829}
  (\bibinfo{year}{2023}).
\newblock


\bibitem[Nickel et~al\mbox{.}(2011)]%
        {nickel2011three}
\bibfield{author}{\bibinfo{person}{Maximilian Nickel}, \bibinfo{person}{Volker
  Tresp}, {and} \bibinfo{person}{Hans{-}Peter Kriegel}.}
  \bibinfo{year}{2011}\natexlab{}.
\newblock \showarticletitle{A Three-Way Model for Collective Learning on
  Multi-Relational Data}. In \bibinfo{booktitle}{\emph{{ICML}}}.
  \bibinfo{publisher}{Omnipress}, \bibinfo{pages}{809--816}.
\newblock


\bibitem[Ristoski and Paulheim(2016)]%
        {Ristoski2016RDF2Vec}
\bibfield{author}{\bibinfo{person}{Petar Ristoski} {and} \bibinfo{person}{Heiko
  Paulheim}.} \bibinfo{year}{2016}\natexlab{}.
\newblock \showarticletitle{RDF2Vec: {RDF} Graph Embeddings for Data Mining}.
  In \bibinfo{booktitle}{\emph{{ISWC}}}, Vol.~\bibinfo{volume}{9981}.
  \bibinfo{pages}{498--514}.
\newblock


\bibitem[Sun et~al\mbox{.}(2019)]%
        {sun2019rotate}
\bibfield{author}{\bibinfo{person}{Zhiqing Sun}, \bibinfo{person}{Zhi{-}Hong
  Deng}, \bibinfo{person}{Jian{-}Yun Nie}, {and} \bibinfo{person}{Jian Tang}.}
  \bibinfo{year}{2019}\natexlab{}.
\newblock \showarticletitle{RotatE: Knowledge Graph Embedding by Relational
  Rotation in Complex Space}. In \bibinfo{booktitle}{\emph{{ICLR} (Poster)}}.
  \bibinfo{publisher}{OpenReview.net}.
\newblock


\bibitem[Sun et~al\mbox{.}(2017)]%
        {sun2017cross}
\bibfield{author}{\bibinfo{person}{Zequn Sun}, \bibinfo{person}{Wei Hu}, {and}
  \bibinfo{person}{Chengkai Li}.} \bibinfo{year}{2017}\natexlab{}.
\newblock \showarticletitle{Cross-Lingual Entity Alignment via Joint
  Attribute-Preserving Embedding}. In \bibinfo{booktitle}{\emph{{ISWC} {(1)}}}
  \emph{(\bibinfo{series}{Lecture Notes in Computer Science},
  Vol.~\bibinfo{volume}{10587})}. \bibinfo{publisher}{Springer},
  \bibinfo{pages}{628--644}.
\newblock


\bibitem[Sun et~al\mbox{.}(2018)]%
        {Sun2018Bootstrapping}
\bibfield{author}{\bibinfo{person}{Zequn Sun}, \bibinfo{person}{Wei Hu},
  \bibinfo{person}{Qingheng Zhang}, {and} \bibinfo{person}{Yuzhong Qu}.}
  \bibinfo{year}{2018}\natexlab{}.
\newblock \showarticletitle{Bootstrapping Entity Alignment with Knowledge Graph
  Embedding}. In \bibinfo{booktitle}{\emph{{IJCAI}}}.
  \bibinfo{publisher}{ijcai.org}, \bibinfo{pages}{4396--4402}.
\newblock


\bibitem[Trisedya et~al\mbox{.}(2019)]%
        {Trisedya2019Entity}
\bibfield{author}{\bibinfo{person}{Bayu~Distiawan Trisedya},
  \bibinfo{person}{Jianzhong Qi}, {and} \bibinfo{person}{Rui Zhang}.}
  \bibinfo{year}{2019}\natexlab{}.
\newblock \showarticletitle{Entity Alignment between Knowledge Graphs Using
  Attribute Embeddings}. In \bibinfo{booktitle}{\emph{{AAAI}}}.
  \bibinfo{publisher}{{AAAI} Press}, \bibinfo{pages}{297--304}.
\newblock


\bibitem[Trouillon et~al\mbox{.}(2016)]%
        {trouillon2016complex}
\bibfield{author}{\bibinfo{person}{Th{\'{e}}o Trouillon},
  \bibinfo{person}{Johannes Welbl}, \bibinfo{person}{Sebastian Riedel},
  \bibinfo{person}{{\'{E}}ric Gaussier}, {and} \bibinfo{person}{Guillaume
  Bouchard}.} \bibinfo{year}{2016}\natexlab{}.
\newblock \showarticletitle{Complex Embeddings for Simple Link Prediction}. In
  \bibinfo{booktitle}{\emph{{ICML}}} \emph{(\bibinfo{series}{{JMLR} Workshop
  and Conference Proceedings}, Vol.~\bibinfo{volume}{48})}.
  \bibinfo{publisher}{JMLR.org}, \bibinfo{pages}{2071--2080}.
\newblock


\bibitem[Vrandecic and Kr{\"{o}}tzsch(2014)]%
        {vrandevcic2014wikidata}
\bibfield{author}{\bibinfo{person}{Denny Vrandecic} {and}
  \bibinfo{person}{Markus Kr{\"{o}}tzsch}.} \bibinfo{year}{2014}\natexlab{}.
\newblock \showarticletitle{Wikidata: a free collaborative knowledgebase}.
\newblock \bibinfo{journal}{\emph{Commun. {ACM}}} \bibinfo{volume}{57},
  \bibinfo{number}{10} (\bibinfo{year}{2014}), \bibinfo{pages}{78--85}.
\newblock


\bibitem[Wang et~al\mbox{.}(2019)]%
        {wang2019learning}
\bibfield{author}{\bibinfo{person}{Meng Wang}, \bibinfo{person}{Haomin Shen},
  \bibinfo{person}{Sen Wang}, \bibinfo{person}{Lina Yao},
  \bibinfo{person}{Yinlin Jiang}, \bibinfo{person}{Guilin Qi}, {and}
  \bibinfo{person}{Yang Chen}.} \bibinfo{year}{2019}\natexlab{}.
\newblock \showarticletitle{Learning to Hash for Efficient Search Over
  Incomplete Knowledge Graphs}. In \bibinfo{booktitle}{\emph{{ICDM}}}.
  \bibinfo{publisher}{{IEEE}}, \bibinfo{pages}{1360--1365}.
\newblock


\bibitem[Wang et~al\mbox{.}(2017)]%
        {Wang2017Knowledge}
\bibfield{author}{\bibinfo{person}{Quan Wang}, \bibinfo{person}{Zhendong Mao},
  \bibinfo{person}{Bin Wang}, {and} \bibinfo{person}{Li Guo}.}
  \bibinfo{year}{2017}\natexlab{}.
\newblock \showarticletitle{Knowledge Graph Embedding: {A} Survey of Approaches
  and Applications}.
\newblock \bibinfo{journal}{\emph{{IEEE} Trans. Knowl. Data Eng.}}
  \bibinfo{volume}{29}, \bibinfo{number}{12} (\bibinfo{year}{2017}),
  \bibinfo{pages}{2724--2743}.
\newblock


\bibitem[Wang et~al\mbox{.}(2018)]%
        {Wang2018GCNAlign}
\bibfield{author}{\bibinfo{person}{Zhichun Wang}, \bibinfo{person}{Qingsong
  Lv}, \bibinfo{person}{Xiaohan Lan}, {and} \bibinfo{person}{Yu Zhang}.}
  \bibinfo{year}{2018}\natexlab{}.
\newblock \showarticletitle{Cross-lingual Knowledge Graph Alignment via Graph
  Convolutional Networks}. In \bibinfo{booktitle}{\emph{{EMNLP}}}.
  \bibinfo{publisher}{Association for Computational Linguistics},
  \bibinfo{pages}{349--357}.
\newblock


\bibitem[Wilkinson et~al\mbox{.}(2016)]%
        {wilkinson2016fair}
\bibfield{author}{\bibinfo{person}{Mark~D Wilkinson}, \bibinfo{person}{Michel
  Dumontier}, \bibinfo{person}{IJsbrand~Jan Aalbersberg},
  \bibinfo{person}{Gabrielle Appleton}, \bibinfo{person}{Myles Axton},
  \bibinfo{person}{Arie Baak}, \bibinfo{person}{Niklas Blomberg},
  \bibinfo{person}{Jan-Willem Boiten}, \bibinfo{person}{Luiz~Bonino da
  Silva~Santos}, \bibinfo{person}{Philip~E Bourne}, {et~al\mbox{.}}}
  \bibinfo{year}{2016}\natexlab{}.
\newblock \showarticletitle{The FAIR Guiding Principles for scientific data
  management and stewardship}.
\newblock \bibinfo{journal}{\emph{Scientific data}} \bibinfo{volume}{3},
  \bibinfo{number}{1} (\bibinfo{year}{2016}), \bibinfo{pages}{1--9}.
\newblock


\bibitem[Yang et~al\mbox{.}(2015)]%
        {yang2015embedding}
\bibfield{author}{\bibinfo{person}{Bishan Yang}, \bibinfo{person}{Wen{-}tau
  Yih}, \bibinfo{person}{Xiaodong He}, \bibinfo{person}{Jianfeng Gao}, {and}
  \bibinfo{person}{Li Deng}.} \bibinfo{year}{2015}\natexlab{}.
\newblock \showarticletitle{Embedding Entities and Relations for Learning and
  Inference in Knowledge Bases}. In \bibinfo{booktitle}{\emph{{ICLR}
  (Poster)}}.
\newblock


\bibitem[Zhang et~al\mbox{.}(2019)]%
        {Zhang2019Multi-view}
\bibfield{author}{\bibinfo{person}{Qingheng Zhang}, \bibinfo{person}{Zequn
  Sun}, \bibinfo{person}{Wei Hu}, \bibinfo{person}{Muhao Chen},
  \bibinfo{person}{Lingbing Guo}, {and} \bibinfo{person}{Yuzhong Qu}.}
  \bibinfo{year}{2019}\natexlab{}.
\newblock \showarticletitle{Multi-view Knowledge Graph Embedding for Entity
  Alignment}. In \bibinfo{booktitle}{\emph{{IJCAI}}}.
  \bibinfo{pages}{5429--5435}.
\newblock


\bibitem[Zhang et~al\mbox{.}(2023)]%
        {zhang2023variational}
\bibfield{author}{\bibinfo{person}{Xiaoyu Zhang}, \bibinfo{person}{Xin Xin},
  \bibinfo{person}{Dongdong Li}, \bibinfo{person}{Wenxuan Liu},
  \bibinfo{person}{Pengjie Ren}, \bibinfo{person}{Zhumin Chen},
  \bibinfo{person}{Jun Ma}, {and} \bibinfo{person}{Zhaochun Ren}.}
  \bibinfo{year}{2023}\natexlab{}.
\newblock \showarticletitle{Variational Reasoning over Incomplete Knowledge
  Graphs for Conversational Recommendation}. In
  \bibinfo{booktitle}{\emph{{WSDM}}}. \bibinfo{publisher}{{ACM}},
  \bibinfo{pages}{231--239}.
\newblock


\bibitem[Zhao et~al\mbox{.}(2022)]%
        {zhao2022improving}
\bibfield{author}{\bibinfo{person}{Fen Zhao}, \bibinfo{person}{Yinguo Li},
  \bibinfo{person}{Jie Hou}, {and} \bibinfo{person}{Ling Bai}.}
  \bibinfo{year}{2022}\natexlab{}.
\newblock \showarticletitle{Improving question answering over incomplete
  knowledge graphs with relation prediction}.
\newblock \bibinfo{journal}{\emph{Neural Comput. Appl.}} \bibinfo{volume}{34},
  \bibinfo{number}{8} (\bibinfo{year}{2022}), \bibinfo{pages}{6331--6348}.
\newblock


\bibitem[Zheng et~al\mbox{.}(2020)]%
        {zheng2020dgl}
\bibfield{author}{\bibinfo{person}{Da Zheng}, \bibinfo{person}{Xiang Song},
  \bibinfo{person}{Chao Ma}, \bibinfo{person}{Zeyuan Tan},
  \bibinfo{person}{Zihao Ye}, \bibinfo{person}{Jin Dong}, \bibinfo{person}{Hao
  Xiong}, \bibinfo{person}{Zheng Zhang}, {and} \bibinfo{person}{George
  Karypis}.} \bibinfo{year}{2020}\natexlab{}.
\newblock \showarticletitle{{DGL-KE:} Training Knowledge Graph Embeddings at
  Scale}. In \bibinfo{booktitle}{\emph{{SIGIR}}}. \bibinfo{publisher}{{ACM}},
  \bibinfo{pages}{739--748}.
\newblock


\end{thebibliography}

\end{document}